\documentclass{article}

\usepackage[preprint]{icml2026}
\usepackage{enumitem}

\usepackage{amsmath}
\usepackage{amssymb}
\usepackage{amsthm}
\usepackage{booktabs}
\usepackage{graphicx}
\usepackage{subcaption}
\usepackage{algorithm}
\usepackage{algcompatible}
\usepackage{makecell}
\usepackage{hyperref}
\usepackage{xcolor}
\usepackage{multirow}
\usepackage{adjustbox}

\newcommand*{\method}{\textsc{Probe-KD}}
\newcommand*{\teacher}{\ensuremath{\mathcal{T}}}
\newcommand*{\student}{\ensuremath{\mathcal{S}}}
\newcommand*{\probe}{\ensuremath{\mathcal{P}}}
\newcommand*{\hidden}{\ensuremath{\mathbf{h}}}
\newcommand*{\softmax}{\ensuremath{\text{softmax}}}

\icmltitlerunning{Task-Specific Knowledge Distillation via Intermediate Probes}

\begin{document}

\twocolumn[
\icmltitle{Task-Specific Knowledge Distillation via Intermediate Probes}

\icmlsetsymbol{equal}{*}
\begin{icmlauthorlist}
    \icmlauthor{Ryan Brown}{equal,oii}
    \icmlauthor{Chris Russell}{oii}
\end{icmlauthorlist}

\icmlaffiliation{oii}{Oxford Internet Institute, University of Oxford, Oxford, United Kingdom}

\icmlcorrespondingauthor{Ryan Brown}{ryan.brown@oii.ox.ac.uk}
\vskip 0.3in
]

\printAffiliationsAndNotice{}

\begin{abstract}

Knowledge distillation from large language models (LLMs) assumes that the teacher's output distribution is a high-quality training signal. On reasoning tasks, this assumption is frequently violated. A model's intermediate representations may encode the correct answer, yet this information is lost or distorted through the vocabulary projection, where prompt formatting and answer-token choices creates brittle, noisy outputs. 

We introduce \method{}, a distillation framework that bypasses this bottleneck by training lightweight probes on frozen teacher hidden states and using the probe's predictions, rather than output logits, as supervision for student training. This simple change yields consistent improvements across four reasoning benchmarks (AQuA-RAT, ARC Easy/Challenge, and MMLU), with gains most pronounced under limited data.

Probes trained on intermediate representations provide cleaner labels than the teacher's own outputs, effectively denoising the distillation signal. \method{} requires no architectural changes to student or teacher, is architecture-agnostic, and adds minimal compute since probe training is cheap and teacher representations can be cached. By exploiting internal representations, \method{} enables practitioners to extract more value from large teacher models without additional training data or architectural complexity.

\end{abstract}

\section{Introduction}
\label{sec:intro}

Large language models (LLMs) have achieved remarkable performance across diverse reasoning tasks, yet deploying these models at scale remains prohibitively expensive. Knowledge distillation~\citep{hinton_distilling_2015} offers a path forward by training compact student models to mimic larger teachers  on specific target domains. The dominant paradigm matches the student's output distribution to the teacher's predicted probabilities~\citep{sanh_distilbert_2020, jiao_tinybert_2020, sun_patient_2019}.

We address the challenge of combining domain-specific annotations alongside model distillation to substantially improve the performance of small BERT-style classifiers on domain-specific benchmarks. Our approach substantially outperforms both standard distillation and classical supervised approaches.

\begin{figure}[t]
    \centering
    \includegraphics[width=\columnwidth]{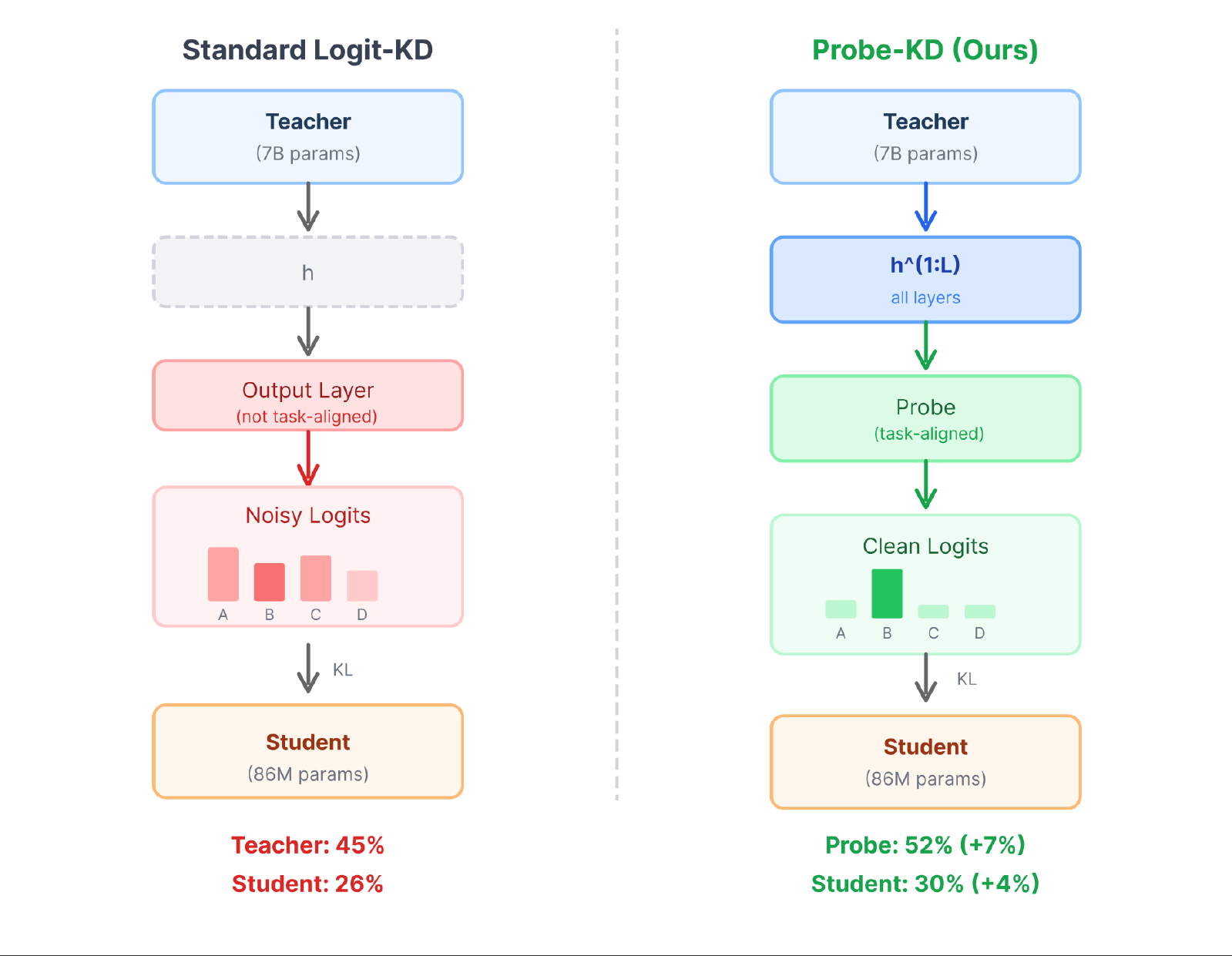}
    \caption{\method{} vs Typical Distillation. In standard logit distillation (left), soft labels come from the teacher's output layer, which projects hidden states onto answer tokens via a fixed, task-agnostic readout. This bottleneck produces noisy supervision even when the correct answer is encoded internally. \method{} (right) bypasses this bottleneck by training a probe to decode hidden states directly, learning a task-aligned projection. On AQuA-RAT, the probe achieves 52\% accuracy versus the teacher's 45\%, demonstrating that cleaner readouts yield cleaner labels which in turn yields more performant students. 
}
    \label{fig:overview}
\end{figure}

\begin{figure}[t]
    \centering
    \includegraphics[width=\columnwidth]{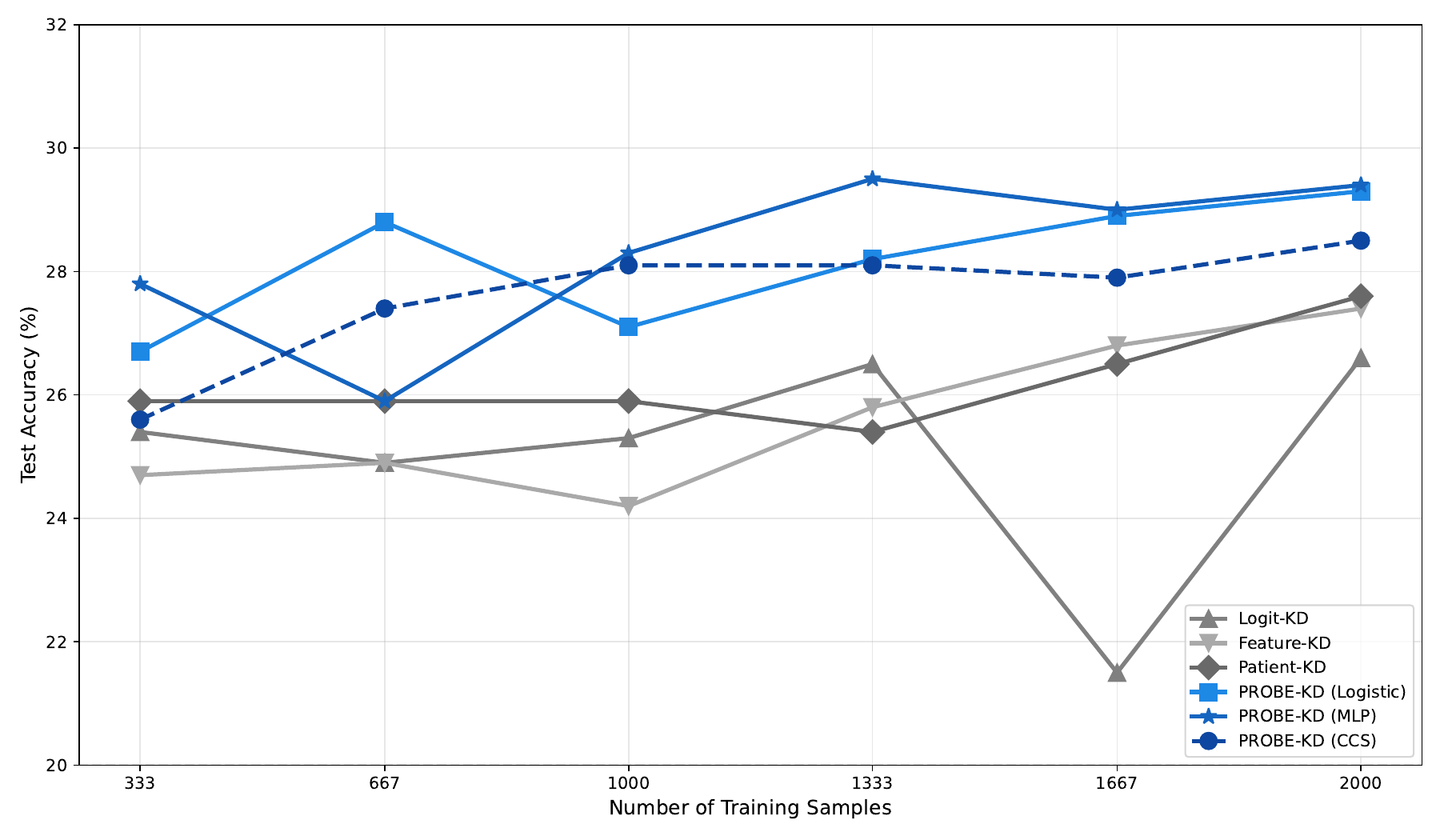}
    \caption{Data efficiency comparison on AQuA-RAT. Test accuracy (\%) as a function of training data percentage. We distill from Qwen2.5-7B-Instruct (teacher) to DeBERTa-v3-base (student, 86M parameters). \method{} variants consistently outperform standard distillation baselines across all data regimes, with gains most pronounced in low-data settings.}
    \label{fig:overview2}
\end{figure}

\begin{table}[t]
    \centering
    \caption{Comparison of knowledge distillation approaches for building smaller task-specific models from LLMs. \method{} is the only approach to be domain adaptable, compute efficient, and to make use of LLMs internal representations.}
    \label{tab:comparison}
    \small
    \begin{tabular}{@{}lccc@{}}
        \toprule
        & \makecell{Domain\\adapted} & \makecell{Compute\\efficient} & \makecell{Uses LLM\\representation} \\
        \midrule
        LLM Distillation      & No  & Yes & Yes \\
        Train on labels only        & Yes & Yes & No  \\
        Fine-tune and distil   & Yes & No  & Yes \\
        \textsc{Probe-KD}      & Yes & Yes & Yes \\
        \bottomrule
    \end{tabular}
\end{table}

The need for additional annotations is clear, as general-purpose reasoners, LLMs often demonstrate limited success on new benchmarks, and when used as teachers, their output becomes a noisy form of supervision. On multiple-choice reasoning tasks, LLMs frequently assign probability mass to incorrect answers, not because their internal representation is insufficiently rich, but because the mapping from internal representations to the correct answer tokens (A, B, C, D) is suboptimal for these benchmarks. Put simply, the teacher's output layer was optimized for general next-token prediction rather than for expressing task-specific knowledge.

Prior work has demonstrated that LLM hidden states encode substantially richer information than their outputs reveal: probes can recover latent knowledge even when models output incorrect answers~\citep{burns_discovering_2024, azaria_internal_2023}, and internal representations encode task-relevant structure that the output layer fails to express~\citep{zou_representation_2025, alain_understanding_2018}.

We exploit this observation for more effective knowledge transfer. Our approach, \method{} (Probe-based Knowledge Distillation), is a simple two-stage procedure (Figure~\ref{fig:overview}). In Stage~1, we extract hidden states from all layers of the teacher for each training example, and train a lightweight MLP probe to predict task labels from these representations. In Stage~2, we freeze the probe, compute its soft predictions for each example, and use these as supervision for training a compact student model via KL-divergence distillation.

As Table~\ref{tab:comparison} summarizes, \method{} is the only approach that is simultaneously domain-adapted, compute-efficient, and leverages LLM representations. Compared to standard distillation, \method{} allows exploitation of additional ground-truth annotations for almost no additional compute. Unlike fine-tuning a model, the most compute intensive component of \method{} is the generation of hidden states, which is a part of generating model outputs for standard distillation. Furthermore, when labels are unavailable, unsupervised probing methods such as CCS \citep{burns_discovering_2024} can replace supervised probe training.

\method{} yields two complementary benefits over supervised learning with hard labels. First, soft probability distributions provide richer supervision per example: beyond indicating which answer is correct, they encode which incorrect answers are plausible, capturing inter-class relationships that improve generalization~\citep{hinton_distilling_2015, muller_when_2020, yuan_revisiting_2021}. \citet{mandal_theoretical_2024} formalize this advantage, proving that soft-label training requires $O(1/\gamma^2)$ neurons to achieve a given loss versus $O(1/\gamma^4)$ for hard labels, a substantial benefit when student capacity is limited. Second, soft labels provide implicit regularization that prevents overconfidence and improves calibration~\citep{zhou_rethinking_2021, szegedy_rethinking_2015}. These benefits compound in our setting, where task-specific data is limited and student capacity is severely constrained.

Compared to standard logit-based distillation, \method{} removes noise from the teacher's output layer. This final projection was never optimized for the downstream task, it maps rich internal representations to arbitrary answer tokens through a decoder trained for next-token prediction. Teacher logits therefore mix useful distributional information with decoder noise. \method{} bypasses this bottleneck by learning a task-specific decoder (the probe) directly on hidden states, producing soft labels that retain dark knowledge while eliminating output-layer artifacts.

While a probe can only decode information already present in the teacher's representations, it can improve task accuracy by exploiting domain specificity. Evidence for this comes from probe accuracy: on AQuA-RAT, an MLP probe achieves 52\% versus 45\% for the teacher's own outputs~\citep{chetelat_innerthoughts_2025}. This gap would be impossible if hidden states did not allow recovery of the correct answer, since the probe has access only to the teacher's internal representation, not the original input. The training labels serve solely to learn the optimal projection from latent space to label space~\citep{alain_understanding_2018}. This aligns with findings that LLMs encode information that they fail to output correctly~\citep{burns_discovering_2024, azaria_internal_2023}; the probe simply provides better task alignment.

Our approach is inherently task-specific, with the probe trained on task data and its outputs reflecting task-relevant structure. This is consistent with evidence that task-specific distillation outperforms task-agnostic approaches~\citep{jiao_tinybert_2020, liu_rethinking_2022}, particularly for compact models where capacity must be allocated carefully. Multiple-choice reasoning is not merely an evaluation format but a common inference primitive: classification, reranking, and decision-support systems all reduce to selecting among constrained options, making compact specialists for this task structure broadly applicable. 

We evaluate \method{} against supervised learning and standard logit distillation on four multiple-choice reasoning benchmarks: AQuA-RAT, ARC-Challenge, ARC-Easy, and MMLU. Here, \method{} achieves state-of-the-art knowledge distillation (Figure~\ref{fig:overview2}). Our contributions include:

\begin{itemize}[leftmargin=*, itemsep=2pt, topsep=2pt]
    \item We introduce \method{}, a distillation framework that fuses domain-specific annotations with LLM internal states via probe predictions. These predictions are used as soft supervision, combining dark knowledge  transfer with task-specific optimization.
    \item We provide a conceptual framework distinguishing \emph{latent information} (contained in the hidden states) from \emph{the teacher's answers} (outputs), showing that distilling the former yields superior students.
    \item \method{} We demonstrate that probe architecture impacts distillation quality. MLP probes consistently outperform linear probes, suggesting sufficient capacity is necessary to decode task-relevant structure from hidden states
\end{itemize}

\section{Related Work}
\label{sec:related}

\textbf{Knowledge Distillation.}
Knowledge distillation trains student networks to match the teacher's soft labels, leveraging the richer supervision they provide over hard labels~\citep{hinton_distilling_2015, bucilua_model_2006}. Extensions include feature-based methods that align intermediate representations~\citep{romero_fitnets_2015}, attention transfer~\citep{zagoruyko_paying_2017}, and relational approaches that preserve pairwise similarities~\citep{park_relational_2019, tian_contrastive_2022}. For language models, DistilBERT~\citep{sanh_distilbert_2020} and TinyBERT~\citep{jiao_tinybert_2020} combine logit matching with hidden-state alignment, while Patient Knowledge Distillation~\citep{sun_patient_2019} distills from multiple intermediate layers. MiniLM~\citep{wang_minilm_2020} transfers self-attention distributions rather than hidden states. Recent work extends these ideas to LLMs: MiniLLM~\citep{gu_minillm_2025} uses reverse KL divergence to avoid overestimating low-probability tokens, while Distilling Step-by-Step~\citep{hsieh_distilling_2023} extracts rationales alongside labels.

Feature-based methods require the student to replicate the teacher's hidden states directly (e.g., minimizing $\|h_{\text{student}} - h_{\text{teacher}}\|^2$). This creates architectural coupling. The student and teacher must share compatible hidden dimensions, and assumes teacher representations are optimal targets for the student. \method{} avoids both issues: we train a probe \emph{on top of} teacher hidden states to produce improved soft labels, then distill from the probe's output distribution. The student never sees teacher hidden states, enabling arbitrary student architectures while leveraging the probe's denoising effect.

\textbf{Chain-of-Thought Distillation.}
A parallel line of work distills reasoning capabilities by training students on teacher-generated rationales~\citep{wei_chain--thought_2023, magister_teaching_2023, shridhar_distilling_2023}. These methods require the teacher to produce explicit chain-of-thought explanations, which the student learns to generate before answering. \citet{deng_implicit_2023} propose implicit chain-of-thought reasoning, distilling reasoning into vertical computation across layers rather than horizontal token generation. \citet{ho_large_2023} show that fine-tuning on LLM-generated rationales can enable small models to outperform few-shot prompted large models. While effective for tasks where rationales are available or can be generated, these approaches do not apply to settings with only answer labels. \method{} operates in the latter regime: we require only hidden states and task labels, making our approach applicable to any classification task without rationale annotation.

\textbf{Probing Neural Networks.}
Linear probes have been extensively used to understand what neural networks encode~\citep{alain_understanding_2018, belinkov_analysis_2019}. For transformers, probing reveals syntactic structure~\citep{hewitt_structural_2019}, semantic roles~\citep{tenney_bert_2019}, and world knowledge~\citep{petroni_language_2019}. \citet{hewitt_designing_2019} introduce control tasks to measure probe selectivity, showing that high probe accuracy does not necessarily indicate information is encoded, powerful probes can learn the task themselves. This concern is less relevant for our setting given that we want the probe to solve the task well, using whatever information the hidden states provide. The question is not whether the probe memorizes versus extracts, but whether probe-derived soft labels improve student training, an empirical question we answer affirmatively.

Recent work extends probing to factual knowledge~\citep{meng_locating_2023} and reasoning~\citep{stolfo_mechanistic_2023}. We build on this tradition but shift the goal. Rather than using probes to \emph{analyze} what models encode, we use them to \emph{extract} improved training signal for distillation.

\textbf{Latent Information in LLMs.}
Several studies demonstrate that LLM hidden states encode richer information than their outputs reveal. \citet{burns_discovering_2024} show that unsupervised probes can recover latent knowledge even when models output incorrect answers, achieving accuracy above zero-shot baselines. \citet{azaria_internal_2023} find that internal states encode whether the model is generating truthful content, independent of the actual output. The logit lens~\citep{nostalgebraist_interpreting_2020} and tuned lens~\citep{belrose_eliciting_2025} decode hidden states at intermediate layers into vocabulary distributions, revealing how predictions are refined across layers. \citet{li_inference-time_2024} show that middle layers often contain the most task-relevant features, not final layers. These works establish that a gap exists between what LLMs encode and what they output. We are the first to exploit this gap for knowledge distillation. Prior work uses probes and lenses for analysis and interpretation; we show that probe predictions provide superior supervision for training compact student models. The tuned lens decodes to the vocabulary space using the model's own unembedding matrix; our probe decodes to task-specific label space using learned projections trained on task data. The teacher's unembedding matrix was optimized for next-token prediction, not for the downstream task, introducing noise that our probe avoids.

\section{Method: Probe-KD}
\label{sec:method}

We now describe \method{} in detail. Let $\teacher$ denote a large teacher model with $L$ layers, and $\student$ a compact student model. Given a dataset $\mathcal{D} = \{(x_i, y_i)\}_{i=1}^N$ of input-label pairs, our goal is to train $\student$ to perform well on the task while being significantly smaller than $\teacher$.

\subsection{Stage 1: Probe Training}
For each input $x$, we extract hidden states from all $L$ layers of $\teacher$ and concatenate them: $\hidden = [\hidden^{(1)}; \hidden^{(2)}; \ldots; \hidden^{(L)}] \in \mathbb{R}^{L \cdot d}$, where $d$ is the hidden dimension and we use the representation at the last token position from each layer.

We train a probe $\probe: \mathbb{R}^{L \cdot d} \rightarrow \mathbb{R}^C$ to predict task labels. We consider two architectures:

\textbf{Linear Probe (Logistic):} $\probe(\hidden) = W\hidden + b$, where $W \in \mathbb{R}^{C \times (L \cdot d)}$.

\textbf{MLP Probe:} $\probe(\hidden) = W_2 \cdot \text{ReLU}(W_1\hidden + b_1) + b_2$, where $W_1 \in \mathbb{R}^{h \times (L \cdot d)}$, $W_2 \in \mathbb{R}^{C \times h}$, and $h$ is a hidden dimension (512 in our experiments).

The probe is trained with cross-entropy loss: $\mathcal{L}_{\text{probe}} = -\sum_{i=1}^N \log \softmax(\probe(\hidden_i))_{y_i}$.

\begin{algorithm}[t]
\caption{\method{}: Probe-Based Knowledge Distillation}
\label{alg:probe_kd}
\begin{algorithmic}[1]
\REQUIRE Teacher $\teacher$ with $L$ layers, Student $\student$, Dataset $\mathcal{D}$
\STATE \textbf{// Stage 1: Extract Hidden States}
\FOR{$(x, y) \in \mathcal{D}$}
    \STATE Extract $\hidden = [\hidden^{(1)}; \ldots; \hidden^{(L)}]$ from $\teacher(x)$
\ENDFOR
\STATE \textbf{// Stage 2: Train Probe}
\IF{supervised}
    \STATE Train probe $\probe$ on $\{(\hidden_i, y_i)\}$ with cross-entropy
\ELSE[\textit{(CCS: unsupervised)}]
    \STATE Train probe $\probe$ on $\{\hidden_i\}$ with consistency + confidence loss
\ENDIF
\STATE \textbf{// Stage 3: Distill to Student}
\FOR{$(x, y) \in \mathcal{D}$}
    \STATE Compute $p_{\text{probe}} = \softmax(\probe(\hidden)/\tau)$
    \STATE Compute $p_{\student} = \softmax(\student(x)/\tau)$
    \STATE $\mathcal{L} = \alpha \cdot \text{KL}(p_{\text{probe}} \| p_{\student}) + (1-\alpha) \cdot \text{CE}(y, \student(x))$
    \STATE Update $\student$ with $\nabla \mathcal{L}$
\ENDFOR
\RETURN Trained student $\student$
\end{algorithmic}
\end{algorithm}

\paragraph{Unsupervised Variant (CCS).}
We also consider an unsupervised probe training approach using Contrast-Consistent Search (CCS)~\cite{burns_discovering_2024}. Unlike the supervised probes which require task labels, CCS discovers a truth direction in the teacher's representation space using only unlabeled contrast pairs. We extend this logic for MCQs by exploiting the fact that one answer must be correct. 

For each question with $C$ choices, we extract hidden states $\hidden_1, \ldots, \hidden_C$ and train a binary probe $\probe_{\text{ccs}}: \mathbb{R}^{L \cdot d} \rightarrow \mathbb{R}$ that predicts whether each choice is correct. Let $p_c = \sigma(\probe_{\text{ccs}}(\hidden_c))$ be the predicted probability for choice $c$. The CCS loss enforces two properties without using labels: (1) confidence predictions should be decisive, not 0.5, via $\mathcal{L}_{\text{conf}} = \frac{1}{C}\sum{c} p_c(1 - p_c)$; and (2) consistency, exactly one answer should be correct, via $\mathcal{L}{\text{cons}} = (\sum{c} p_c - 1)^2$. The probe is optimized with $\mathcal{L}{\text{ccs}} = \mathcal{L}{\text{conf}} + \mathcal{L}_{\text{cons}}$, requiring no labeled examples.

\subsection{Stage 2: Probe-Based Distillation}

Once the probe is trained, we use its soft predictions as supervision for the student. Given input $x$, we compute the probe's output distribution:
\begin{equation}
    p_{\text{probe}} = \softmax(\probe(\hidden)/\tau)
\end{equation}
where $\tau$ is the temperature parameter and $\hidden$ is the concatenated all-layer representation.

The student is trained with a combination of distillation and task losses:
\begin{equation}
    \mathcal{L} = \alpha \cdot \text{KL}(p_{\text{probe}} \| p_{\student}) + (1-\alpha) \cdot \text{CE}(y, \student(x))
\end{equation}
where $\alpha$ balances the two objectives.

Algorithm~\ref{alg:probe_kd} summarizes the complete \method{} procedure.

\subsection{Why Probes Provide Better Supervision}

Compared to standard distillation, the use of probes has several distinct advantages. Not only does
training the probe on ground-truth data correct errors made by the existing model, the probe also serves as an intermediary that directly matches the output form of the student. 

In contrast, when using the general LLM model as a teacher, the outputs are always overcomplete. The model must express its ``belief'' through specific tokens, competing with many alternatives most of which cannot be a valid answer. However, a probe learns the optimal mapping from hidden states to labels using ground-truth supervision. This creates a new form of `dark knowledge' \citep{hinton_distilling_2015}, which better reflects the ground-truth while providing an interface to the teacher's internal representations.

\section{Experimental Setup}
\label{sec:experiments}

\subsection{Datasets}

We evaluate on four multiple-choice reasoning benchmarks. AQuA-RAT \citep{ling_program_2017} contains algebraic word problems with 5 answer choices. ARC-Challenge and ARC-Easy \citep{clark_think_2018} contain grade-school science questions with 4 choices; we use the official train/test splits (1.1K/1.2K and 2.3K/2.4K respectively). MMLU \citep{hendrycks_benchmarking_2019} spans 57 subjects across STEM, humanities, and social sciences with 4 choices; since MMLU lacks a standard training set, we use the auxiliary train split provided by HuggingFace (2.2K examples). To study data efficiency, we train on \{1\%, 10\%, 25\%, 50\%, 75\%, 100\%\} subsets of each training set. 

\paragraph{Teacher (Primary).} Qwen2.5-7B-Instruct \citep{qwen_qwen25_2025} (7B parameters, 28 layers, hidden dim 3584). We evaluate via 5-shot multiple-choice prompting.

\paragraph{Teacher Family Ablation.} To test generalization across architectures, we also evaluate Phi-3-mini-4k-instruct (3.8B) and TinyLlama-1.1B-Chat-v1.0 (1.1B) as teachers.

\paragraph{Student (Primary).} DeBERTa-v3-base \citep{he_debertav3_2023} (86M parameters) with a classification head over [CLS] representations.

\paragraph{Student Architecture Ablation.} We compare four student architectures: DeBERTa-v3-base (86M), DeBERTa-v3-large (304M), ModernBERT-base (149M), and ModernBERT-large (395M).

\paragraph{Probes.} We compare three probe architectures trained on concatenated hidden states from all $L$ teacher layers. The \textbf{Logistic} probe is a linear projection $W \in \mathbb{R}^{C \times (L \cdot d)}$. The \textbf{MLP} probe is a two-layer network with hidden dimension 512. The \textbf{CCS} probe~\citep{burns_discovering_2024} is an unsupervised two-layer network that outputs a scalar score per choice, trained without labels. For MCQA, we adapt CCS by treating the $C$ answer choices as the contrasting set: we extract hidden states for each ``Question: \{q\} Answer: \{choice\}'' prompt and train the probe to satisfy $\sum_{c=1}^{C} p_c \approx 1$ (exactly one answer is correct) while maximizing confidence via $\mathcal{L}_{\text{CCS}} = \text{Var}(p) + (\sum_c p_c - 1)^2$, where $p_c = \sigma(f(h_c))$ is the probe's prediction for choice $c$.
\subsection{Methods Compared}

We evaluate seven methods. As baselines, we include the untrained Student-Base and Teacher-MC (teacher 5-shot accuracy). For student training, we compare Supervised (gold labels only), Logit-KD (distillation from teacher output probabilities), and Feature-KD (student hidden states trained to match teacher hidden states via MSE loss \citep{jiao_tinybert_2020}). Our proposed methods are Probe-KD (Logistic) and Probe-KD (MLP), which distill from probe predictions rather than teacher outputs.

\subsection{Training Details}

\emph{Probe Training.} Probes are trained for 20 epochs with AdamW (lr=1e-3, batch size 128, weight decay 0.01). We extract and concatenate hidden states from all  layers (e.g., $28 \times 3584 = 100,352$ dimensions for Qwen2.5-7B).

\emph{Student Training.} 3 epochs, AdamW (lr=2e-5), batch size 16, linear warmup (10\% steps). For distillation: temperature $\tau=2.0$, KD weight $\alpha=0.7$.

\emph{Data Scaling.} We train on \{1\%, 10\%, 25\%, 50\%, 75\%, 100\%\} of training data to study data efficiency.

\emph{Seeds.} Main experiments use 5 seeds (42--46); ablations use 3 seeds.

\section{Results}
\label{sec:results}

We evaluate \method{} against supervised learning, standard distillation methods, and additional baselines across four reasoning benchmarks. Our experiments address three questions: (1) Does \method{} improve over standard distillation? (2) When are the gains largest? (3) Does the approach generalize across teachers, students, and domains?

\subsection{Main Results}

Table~\ref{tab:probe_quality} establishes that MLP probes trained on hidden-states consistently outperform the teacher's own outputs. The gap is non-trivial, up to +5.6\% on AQuA-RAT, confirming that hidden states encode task-relevant information that the output layer fails to express. Linear probes underperform on some datasets, indicating that sufficient probe capacity might be necessary to decode this latent signal. 

\begin{table}[t]
\centering
\caption{Probe accuracy vs.\ teacher accuracy (\%). MLP probes on hidden states outperform teacher 5-shot outputs, with the largest gap on AQuA-RAT.}
\label{tab:probe_quality}
\vskip 0.1in
\begin{tabular}{lcccc}
\toprule
Dataset & Teacher & Logistic & MLP & $\Delta$ \\
\midrule
AQuA-RAT & 44.7 & 50.6 & 50.3 & +5.6 \\
ARC-Easy & 96.6 & 96.6 & 97.2 & +0.6 \\
ARC-Challenge & 89.7 & 90.3 & 91.2 & +1.5 \\
MMLU & 71.4 & 70.1 & 73.5 & +2.1 \\
\bottomrule
\end{tabular}
\end{table}

Table~\ref{tab:main_results} presents our main comparison across training methods. \method{} (MLP) achieves the best student accuracy on AQuA (29.4\%) and ARC-E (75.1\%), and outperforms all distillation baselines on average: +2.2\% over Logit-KD, +5.0\% over Feature-KD, and +1.5\% over Patient-KD. 

\begin{table}[t]
\centering
\caption{Test accuracy (\%).}
\label{tab:main_results}
\vskip 0.1in
\setlength{\tabcolsep}{3pt} 
\begin{adjustbox}{width=\columnwidth}
\begin{tabular}{lcccc}
\toprule
Method & AQuA & ARC-C & ARC-E & MMLU \\
\midrule
\multicolumn{5}{l}{\textit{Reference}} \\
Teacher 5-shot & 44.7 & 89.7 & 96.6 & 71.4 \\
MLP Probe & 50.3 & 91.2 & 97.2 & 73.5 \\
\midrule
\multicolumn{5}{l}{\textit{Student: DeBERTa-v3-base}} \\
\multicolumn{5}{l}{\textit{(86M params)}} \\ 
Supervised & 29.3 & 52.3 & 73.5 & 31.8 \\
+ Label Smoothing & 27.6 & 51.6 & 73.9 & 33.8 \\
\hspace{3mm}\citep{szegedy_rethinking_2015}\\
\midrule
\multicolumn{5}{l}{\textit{Distillation Methods}} \\
Logit-KD~\citep{hinton_distilling_2015} & 26.6 & 50.9 & 74.4 & 24.5 \\
Feature-KD~\citep{jiao_tinybert_2020} & 27.4 & 38.6 & 69.7 & 29.6 \\
Patient-KD~\citep{sun_patient_2019} & 27.6 & \textbf{51.5} & 74.6 & 25.7 \\
\midrule
\multicolumn{5}{l}{\textit{Ours}} \\
\method{} (Logistic) & 29.3 & \textbf{51.5} & 74.3 & 27.1 \\
\method{} (MLP) & \textbf{29.4} & 50.1 & \textbf{75.1} & \textbf{30.7} \\
\method{} (CCS) & 28.5 & 49.7 & 74.1 & 26.8 \\
\bottomrule
\end{tabular}
\end{adjustbox}
\end{table}

\begin{figure}[t]
    \includegraphics[width=\columnwidth]{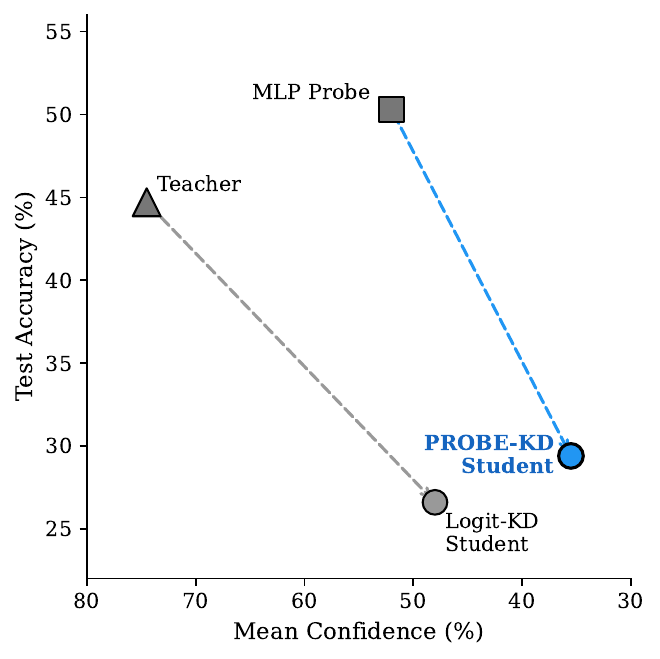}
    \caption{\method{} Calibration analysis on AQuA-RAT. We plot test accuracy against mean prediction confidence, with the x-axis reversed so that lower confidence (better calibration) appears rightward. Arrows indicate the distillation path from source (Teacher or MLP Probe) to student.}
    \label{fig:calibration}
\end{figure}

Several comparisons merit discussion:

\textbf{Feature-KD vs.\ \method{}.} Feature-based distillation~\citep{jiao_tinybert_2020} trains the student to match teacher hidden states directly via MSE loss. This improves over Logit-KD but underperforms \method{}. We attribute this gap to two factors: (1) Feature-KD requires architectural compatibility (we use a projection layer to match dimensions, following \citet{jiao_tinybert_2020}), while \method{} places no constraints on student architecture; (2) Feature-KD treats all information in hidden states as equally valuable, while the probe learns to extract task-relevant signal.

\textbf{Label smoothing.} Uniform label smoothing~\citep{szegedy_rethinking_2015} improves over hard-label supervision but substantially underperforms \method{}. Label smoothing provides uniform regularization; probe soft labels provide informed uncertainty based on the teacher's representations.

\textbf{Teacher fine-tuning.} A natural question arises: if probe training requires labels, why not simply fine-tune the teacher? Table~\ref{tab:ft_comparison} shows these approaches are complementary on AQuA-RAT. We fine-tune the Qwen2.5-7B teacher using LoRA~\citep{hu_lora_2021} on the training set for 3 epochs, improving its 5-shot accuracy from 44.7\% to 52.3\%. However, standard Logit-KD from this fine-tuned teacher yields only 27.8\% student accuracy---barely better than distilling from the base teacher (26.6\%). 

\begin{table}[t]
\centering
\caption{Fine-tuning vs.\ probe training on AQuA-RAT. The teacher is fine-tuned with LoRA for 3 epochs. \method{} improves distillation from both base and fine-tuned teachers.}
\label{tab:ft_comparison}
\vskip 0.1in
\begin{tabular}{lcc}
\toprule
Method & Teacher Acc. & Student Acc. \\
\midrule
\multicolumn{3}{l}{\textit{Base Teacher (Qwen2.5-7B, no fine-tuning)}} \\
Logit-KD & 44.7 & 26.6 \\
\method{} (MLP) & 44.7 & \textbf{29.4} \\
\method{} (CCS)$^\dagger$ & 44.7 & 28.5 \\
\midrule
\multicolumn{3}{l}{\textit{Fine-tuned Teacher (+LoRA, 3 epochs)}} \\
Logit-KD & 52.3 & 27.8 \\
\method{} (MLP) & 52.3 & 28.7 \\
\bottomrule
\multicolumn{3}{l}{\scriptsize $^\dagger$CCS requires no labels for probe training.}
\end{tabular}
\end{table}

In contrast, training an MLP probe on the \emph{base} teacher's hidden states achieves 29.4\%, surpassing the fine-tuned teacher approach. Crucially, \method{} can also be applied \emph{on top of} fine-tuning: training a probe on the fine-tuned teacher's hidden states yields 28.7\%. This suggests that probe-based distillation provides distinct benefits, extracting knowledge that fine-tuning alone cannot transfer through standard distillation.

Moreover, probe training is substantially more efficient: our 53M-parameter MLP probe trains in under 5 minutes on cached hidden states, compared to a few hours for LoRA fine-tuning a 7B model on a B200 GPU, a $>$35$\times$ speedup. 

\subsection{When Does \method{} Help Most?}

We hypothesize that \method{} provides the largest gains when teacher outputs are noisiest. We test this along data availability.

\textbf{Data availability.} Table~\ref{tab:data_scaling} varies training set size on AQuA-RAT. The gap between \method{} and Logit-KD is largest in the lowest data regime and decreases slightly at thousands of samples. The gap over Feature-KD follows a similar pattern. Cleaner supervision from probes is particularly valuable when training examples are scarce, as each example must carry more signal.

\begin{table}[t]
\centering
\caption{Data scaling on AQuA-RAT. \method{} consistently outperforms baseline distillation methods across data regimes.}
\label{tab:data_scaling}
\vskip 0.1in
\begin{tabular}{lccc}
\toprule
 & \multicolumn{3}{c}{Training Samples} \\
\cmidrule(lr){2-4}
Method & $n$=333 & $n$=667 & $n$=1000 \\
\midrule
\multicolumn{4}{l}{\textit{Baseline Distillation}} \\
Logit-KD & 25.4 & 24.9 & 25.3 \\
Feature-KD & 24.7 & 24.9 & 24.2 \\
Patient-KD & 25.9 & 25.9 & 25.9 \\
\midrule
\multicolumn{4}{l}{\textit{Ours}} \\
\method{} (Logistic) & 26.7 & \textbf{28.8} & 27.1 \\
\method{} (MLP) & \textbf{27.8} & 25.9 & \textbf{28.3} \\
\method{} (CCS)$^\dagger$ & 25.6 & 27.4 & 28.1 \\
\midrule
\midrule
 & \multicolumn{3}{c}{Training Samples} \\
\cmidrule(lr){2-4}
Method & $n$=1333 & $n$=1667 & $n$=2000 \\
\midrule
\multicolumn{4}{l}{\textit{Baseline Distillation}} \\
Logit-KD & 26.5 & 21.5 & 26.6 \\
Feature-KD & 25.8 & 26.8 & 27.4 \\
Patient-KD & 25.4 & 26.5 & 27.6 \\
\midrule
\multicolumn{4}{l}{\textit{Ours}} \\
\method{} (Logistic) & 28.2 & 28.9 & 29.3 \\
\method{} (MLP) & \textbf{29.5} & \textbf{29.0} & \textbf{29.4} \\
\method{} (CCS)$^\dagger$ & 27.5 & 27.9 & 28.5 \\
\bottomrule
\multicolumn{4}{l}{\scriptsize $^\dagger$CCS requires no labels for probe training.}
\end{tabular}
\end{table}

\subsection{Ablations}

We verify that \method{} generalizes across model choices.

\textbf{Teacher architecture.} Table~\ref{tab:teacher_ablation} compares two additional teacher families on AQuA-RAT.

\begin{table}[t]
\centering
\caption{Teacher ablation on AQuA-RAT. .}
\label{tab:teacher_ablation}
\vskip 0.1in
\begin{tabular}{lcccc}
\toprule
Teacher & Logit & Feature & \shortstack{\method{} \\ (MLP)}  \\
\midrule
TinyLlama-1.1B & 22.2 & {27.1} & \textbf{27.5}  \\
Phi-3-mini-3.8B & 22.2 & 27.0 & \textbf{27.8} \\
\bottomrule
\end{tabular}
\end{table}

\textbf{Student architecture.} Table~\ref{tab:student_ablation} varies student capacity. \method{} benefits all configurations, with larger gains for larger students (+4.2\% for 86M params vs.\ +5.6\% for 395M over Logit-KD), suggesting that higher-capacity students better exploit the cleaner supervision signal. The gap over Feature-KD is also consistent across architectures.

\begin{table}[t]
\centering
\caption{Student ablation on AQuA-RAT.}
\label{tab:student_ablation}
\vskip 0.1in
\begin{tabular}{lccc}
\toprule
\textbf{DeBERTa} & Logit & Feat. & \method{} \\
\midrule
Base (86M) & 27.8 & 27.6 & \textbf{28.7} \\
Large (304M) & \textbf{21.6} & 20.2 & 21.3 \\
\midrule
\textbf{ModernBERT} & Logit & Feat. & \method{} \\
\midrule
Base (149M) & 27.4 & 26.8 & \textbf{30.2} \\
Large (395M) & 27.9 & 27.4 & \textbf{28.6} \\
\bottomrule
\end{tabular}
\end{table}

\subsection{Calibration}

Beyond accuracy, we examine whether \method{} produces better-calibrated students. We measure calibration via mean confidence: the average of $\max_c p(c \mid x)$ across test examples, where $p(c \mid x)$ is the model's predicted probability for choice $c$. A well-calibrated model has mean confidence approximately equal to its accuracy.

Figure~\ref{fig:calibration} reveals the the teacher LLM is severely overconfident: 74.5\% mean confidence despite only 44.7\% accuracy, a calibration gap of nearly 30 percentage points. Standard Logit-KD transfers this miscalibration to the student, which predicts with 48\% confidence while achieving only 26.6\% accuracy.

In contrast, the MLP probe trained on hidden states is well-calibrated (52\% confidence, 50.3\% accuracy), and the \method{} student inherits this property: its 35.5\% mean confidence closely matches its 29.4\% accuracy. This calibration improvement arises because probes produce soft labels that reflect genuine uncertainty in the hidden representations, rather than the teacher's overconfident token probabilities. Calibrated predictions are particularly valuable in downstream applications where prediction confidence informs decision-making, such as selective prediction.

\subsection{Overview}

Notably, \method{} is the only distillation method that consistently matches or exceeds fully supervised learning across all four benchmarks, a result that standard distillation methods fail to achieve. This suggests that probe-based supervision successfully integrates the benefits of soft-label training with task-specific optimization, yielding students that inherit the teacher's dark knowledge.

\section{Limitations}
\label{sec:limitations}

\textbf{Task scope.} We evaluate on multiple-choice classification, where the probe maps hidden states to a small label set. Extending to generation tasks (e.g., open-ended QA, summarization) would require probes that decode to sequences, substantially increasing complexity. The multiple-choice setting is practically important---it underlies classification, reranking, and constrained generation, but does not cover all distillation scenarios.

\textbf{Hidden state storage.} While probe training is fast (minutes on cached hidden states), extracting and storing hidden states from all layers requires substantial memory: $O(N \cdot L \cdot d)$ for $N$ examples, $L$ layers, and hidden dimension $d$. For Qwen2.5-7B with 28 layers and $d=3584$, this is $\sim$400KB per example. For large datasets, streaming or dimensionality reduction may be necessary.

\textbf{Access requirements.} \method{} requires access to teacher hidden states, precluding black-box API-only teachers. This is a meaningful constraint as many powerful models (e,g. GPT-5.2) do not expose internal representations. However, the proliferation of open-weight models (Llama, Qwen, Mistral) makes this assumption practical.

\textbf{Probe architecture.} We compare linear and two-layer MLP probes; the optimal architecture may vary by task and teacher. More expressive probes could extract additional signal but risk overfitting. We leave systematic architecture search to future work.

\section{Conclusion}
\label{sec:conclusion}

We introduced \method{}, a knowledge distillation framework that uses probe predictions on teacher hidden states as soft supervision. The key insight is that LLM hidden states encode richer task-relevant information than outputs reveal. By training a lightweight probe to decode this latent information, we obtain accurate soft labels that improve student training over standard logit distillation.

Probes trained on hidden states outperform teacher outputs (Table~\ref{tab:probe_quality}), and this improved supervision translates to better students (Table~\ref{tab:main_results}). \method{} achieves state-of-the-art performance across difficult reasoning tasks and in low-data regimes (Table~\ref{tab:data_scaling}). \method{} is consistent across teacher architectures, student architectures, and domains. This suggests distillation should target classifier outputs based on the latent space, rather than the output of the unembedding layer. For tasks where: teacher outputs are unreliable; multi-step reasoning; out-of-distribution inputs; tasks far from pretraining, \method{} offers a principled way to extract cleaner supervision.

\method{} requires no architectural changes to teacher or student, adds minimal overhead, and integrates with any soft-label distillation objective. We hope this work encourages further investigation into representation-based knowledge transfer, moving beyond the assumption that model outputs are the best available supervision signal.


\section*{Impact Statement}
This paper presents work whose goal is to advance the field of machine learning. By showing how to effectively distill expensive LLMs while maintaining a high level of performance we provide a path to reducing some of the environmental impacts of LLM at inference time.

\bibliography{references}
\bibliographystyle{icml2026}

\end{document}